%% file: main.tex
\documentclass[twoside]{article}

\usepackage[accepted]{aistats2024}

\usepackage[round]{natbib}
\renewcommand{\bibname}{References}
\renewcommand{\bibsection}{\subsubsection*{\bibname}}
\newcommand{\xmark}{$\times$}%

\usepackage[utf8]{inputenc}
\input{math_commands.tex}
\usepackage{hyperref}
\usepackage{url}
\usepackage[utf8]{inputenc} 
\usepackage[T1]{fontenc}    
\usepackage{booktabs}       
\usepackage{nicefrac}       
\usepackage{microtype}      
\usepackage{graphicx}
\usepackage{comment}
\usepackage{wrapfig}
\usepackage{amsthm}
\usepackage{booktabs}
\usepackage{balance}
\input{ide_preamble}

\newcommand{\ours}{\texttt{ISAHP}}

\input{math_commands.tex}

\begin{document}
\runningtitle{Self-attentive Hawkes Processes}
\twocolumn[

\aistatstitle{Learning Granger Causality from Instance-wise \\ Self-attentive Hawkes Processes}

\aistatsauthor{ Dongxia Wu$\dagger$\thanks{Work done during internship at IBM} \And Tsuyoshi Id\'e$^*$ \And  Aur\'elie Lozano$^*$ \And  Georgios Kollias$^*$ \And Ji\v{r}\'i Navr\'{a}til$^*$ }
\aistatsauthor{ Naoki Abe$^*$ \And Yi-An Ma$\dagger$ \And Rose Yu$\dagger$}

\aistatsaddress{ $\dagger$University of California, San Diego, \qquad
$^*$IBM T. J. Watson Research Center\\ \texttt{\{dowu,yianma,roseyu\}@ucsd.edu}, \qquad \texttt{\{tide,aclozano,gkollias,jiri,nabe\}@us.ibm.com}} ]

\begin{abstract}
We address the problem of learning Granger causality from asynchronous, interdependent, multi-type event sequences. In particular, we are interested in discovering \textit{instance-level} causal structures in an unsupervised manner.  Instance-level causality identifies causal relationships among individual events, providing more fine-grained information for decision-making.  Existing work in the literature either requires strong assumptions, such as linearity in the intensity function, or heuristically defined model parameters that do not necessarily meet the requirements of Granger causality. 
We propose Instance-wise Self-Attentive Hawkes Processes (\ours{}), a novel deep learning framework that can directly infer the Granger causality at the event instance level. 
\ours{} is the first neural point process model that meets the requirements of Granger causality. It leverages the self-attention mechanism of the transformer to align with the principles of Granger causality.
We empirically demonstrate that \ours{} is capable of discovering complex instance-level causal structures that cannot be handled by classical models. We also show that \ours{} achieves state-of-the-art performance in proxy tasks involving type-level causal discovery and instance-level event type prediction.  
\end{abstract}

\section{Introduction}\label{sec:introduction}

\input{sections/intro}
\section{Background}\label{sec:Background}
\input{sections/background}

\section{Related Work}

\input{sections/related}
\section{Instance-wise Self-attentive Hawkes Processes}\label{sec:Methodology}
\input{sections/method_new}

\section{Experiments}\label{sec:experiments}
\input{sections/exp}

\section{Concluding Remarks}
\input{sections/conc}

\subsubsection*{Acknowledgments}

This work was conducted as an internship project of Dongxia Wu at IBM Research. Yi-An Ma and Rose Yu are partially supported by the U.S. Army Research Office under Army-ECASE award W911NF-07-R-0003-03, the U.S. Department Of Energy, Office of Science, IARPA HAYSTAC Program, CDC-RFA-FT-23-0069, NSF Grants \#2205093, \#2146343, \#2134274, SCALE MoDL-2134209, and CCF-2112665 (TILOS).

\balance
\bibliography{iclr2023_conference}
\bibliographystyle{apalike}

\onecolumn
\section{SUPPLEMENTARY MATERIAL}
\input{sections/appendix}

\end{document}

%% file: math_commands.tex
\usepackage{amsmath,amsfonts,bm}

\def\eqref#1{equation~\ref{#1}}

\def\1{\bm{1}}

\DeclareMathAlphabet{\mathsfit}{\encodingdefault}{\sfdefault}{m}{sl}
\SetMathAlphabet{\mathsfit}{bold}{\encodingdefault}{\sfdefault}{bx}{n}

%% file: ide_preamble.tex
\usepackage{algorithm}
\usepackage{algorithmic}
\usepackage{wrapfig}
\usepackage{graphicx}
\usepackage{mathtools} 
\usepackage{amssymb}
\usepackage{amsmath}
\usepackage{amsthm} 
\usepackage{color}
\usepackage{bm}

\usepackage{tcolorbox}
\usepackage{url}

\newcommand{\bmh}{\bm{h}}

\newcommand{\bmv}{\bm{v}}
\newcommand{\bmw}{\bm{w}}
\newcommand{\bmx}{\bm{x}}

\newcommand{\rmd}{\mathrm{d}}
\newcommand{\rme}{\mathrm{e}}

\newcommand{\sfA}{\mathsf{A}}

\newcommand{\sfK}{\mathsf{K}}

\newcommand{\sfV}{\mathsf{V}}
\newcommand{\sfW}{\mathsf{W}}
\newcommand{\sfX}{\mathsf{X}}

\newcommand{\calD}{\mathcal{D}}

\newcommand{\calH}{\mathcal{H}}

%% file: sections/intro.tex
Automated causal discovery from noisy time-series data is a fundamental problem for machine learning in complex domains. Granger causality~\citep{Granger69Econometrica} is a popular notion of (pseudo) causality in time series data. While extensive prior work exists on Granger causality from regular time series, with the vector autoregressive model being a primary tool (see, e.g.,
~\citep{shojaie2022granger} for the latest review),
relatively less work has been done on analogous problems on discrete {\em event} sequences. These event sequences often occur at irregular intervals. There could also be multiple types of events interacting with each other. Together, they pose significant challenges to Granger causality structure learning  tasks.

Recently, many have used \textit{point process} model as a vehicle for causal discovery of stochastic temporal events~\citep{xu2016learning,eichler2017graphical,zhang2020cause}. 
In particular, Hawkes process~\citep{hawkes1971spectra,hawkes2021personal} is one of the basic building blocks for Granger-causal analysis of \textit{multi-type  event sequences} (i.e., event data that come with a type attribute, indicating what the event is). Thanks to its additive structure over the historical events in the event intensity function, the task of causal discovery can be performed by maximum likelihood estimation of the kernel matrix, which characterizes the causal relationship among different event types. 

However, causal structure among event types only provides coarse-grained information that is aggregated over the event sequence. It lacks fine-grained details of causal relationships among individual events. 
From a practical perspective,  extracting instance-level information for causal analysis is critical. In medical diagnosis, for example, we are interested in capturing precursor events that may have caused a specific symptom of patients. These precursors would be valuable for early detection, screening, and treatment as a prevention. Aggregated type-level causality would only reveal information about generic symptoms categories, rather than identifying the exact precursors that progress directly into diseases. 

For instance-level causal analysis, there are three lines of research to date. One is based on the classical Hawkes process, typically through the minorization-maximization (MM) framework~\citep{veen2008estimation,lewis2011nonparametric,li2015energy,wang2017human,Ide21NeurIPS}. But a classical Hawkes model places the linearity assumption in the intensity function, limiting its expressive power.  Neural point process models ~\citep{xiao2019learning,zuo2020transformer,zhang2020self} are designed to address this limitation but they typically embed event history in the form of a latent state vector losing instance-wise information. As a result, the attention-based score does not necessarily represent the Granger causality.  The third one is based on a post-processing step following maximum likelihood ~\citep{zhang2020cause}, which incurs additional computational costs. 

In this paper, we propose a novel deep Hawkes process model, the ``instance-wise self-attentive Hawkes process (\ours{}),'' that achieves better expressiveness while also enabling direct instance-level Granger-causal analysis. From a mathematical perspective, the key design principle is to maintain an \textit{additive structure}, where causal interaction is represented as the summation of individual historical events. To capture complex causal interactions potentially involving multiple events, we leverage the self-attention mechanism of the Transformer model~\citep{vaswani2017attention}. \ours{} can directly capture instance-wise causal relationships with its additive structure. We can also easily obtain type-level causal relationships by simple aggregation. To the best of our knowledge, \ours{} is the first deep point process model that allows direct instance-wise causal analysis without 
post-processing. 

We empirically demonstrate that \ours{} can discover complex instance-level causal structures that cannot be handled by the classical models and neural point process models without post-processing. Furthermore, our experiments show that \ours{} achieves state-of-the-art performance in two proxy tasks, one involving type-level causal discovery and the other involving instance-level event type prediction. It confirms that the instance- and type-level causal inference tasks are coupled, and our proposed framework manages to model them coherently and holistically.


%% file: sections/background.tex
\subsection{Type-Level Granger Causality}

We are given a training data set $\calD$ with $S$ event sequences, each of which contains $L_s$ events.
\begin{align}
\calD \triangleq
    \{(t^s_i,k^s_i) \mid i \in \{1,\ldots, L_s\}, s \in \{1,\ldots,S\}\}.
\end{align}
In the data set, each event is represented by its timestamp of occurrence $t$ and a type attribute $k$. The timestamps are sorted so that $t^s_i \geq t^s_j$ for $i > j$. We assume that the total number of event types is $K$ and therefore $k^s_i \in \{1,\ldots,K\}$.

We formalize the problem of causal discovery from event sequences as a \textit{unsupervised density estimation} task. Given the history of events $\calH_t=\{(t^s_i,k^s_i)_{t_i<t}\}$, temporal point processes are generally characterized by a conditional distribution called the intensity function $\lambda_k(t\mid \calH_t)$. The intensity function describes the expected rate of occurrence for event type $k$ at a future time point $t$,  and is assumed to have a specific parametric form for causal discovery. For the classical multivariate Hawkes process (MHP), we assume a simple linear form.
\begin{align}
\lambda_k(t \mid \mathcal{H}_t) = \mu_k + 
\sum_{i: t_i < t} \alpha_{k,k_i} \phi_{k,k_i}(t-t_i)
\label{eq:mhp_intensity}
\end{align}
where $\mu_{k}$ is the background intensity for event type $k$, $\{\alpha_{k,k_i}\}$ forms a ${K\times K}$ matrix called the kernel matrix representing the type-level causal influence, and $\phi_{k,k_i}(\cdot)$ is the decay function 
of the causal influence. 

The additive form in Eq.~(\ref{eq:mhp_intensity}) naturally leads to causal interpretation  among event types. For example, if $\alpha_{1,2}=0$, the probability of the next event occurrence of type-1 is not affected at all by the type-2 events in the history. This is indeed the definition of Granger-\textit{non}-causality in point processes~\citep{xu2016learning}. In general, event A is said to be Granger-non-cause of another event B if A does not affect the occurrence probability of event B. 

\subsection{Instance-Level Granger Causality}

Although MHP in Eq.~(\ref{eq:mhp_intensity})  provides  Granger causality interpretations for event sequences, such a causality structure is only at the  \textit{type-level}, instead of \textit{instance-level}. 
To obtain instance-level causality,  a direct generalization of MHP  for an event $i$ with event type $k$ would be
\begin{equation}
\lambda_{i,k}(t \mid \mathcal{H}_t) = \mu_{i,k} + 
\sum_{ j: t_j<t} \alpha_{i,j,k} \phi_{i,j,k}(t-t_j)
    \label{eq:mhp_instance}
\end{equation}
where $\mu_{i,k}$ is the background intensity for event $i$ with event type $k$, $\{\alpha_{i,j,k}\}$ forms a $L\times L\times K$ tensor representing instance-level Granger causality, and $\phi_{i,j,k}(\cdot)$ is the decay function representing time-decay of the causal influence. We assume a maximum sequence length $L=\max(\{L_s\})$ and use padding for varying length sequences.


For instance-level causal analysis, the unsupervised causal discovery problem can be reduced to fitting the model parameters contained in the intensity function $\lambda_{i,k}(t\mid \calH_t)$. However,
such an analysis is very challenging. First, capturing long-range causality between events can be difficult due to the intricate nature of temporal dependencies. Second, it requires a substantial number of parameters to adequately parameterize the model, increasing the risk of overfitting.
In our model, we use a particular parametric form for $\lambda_{i,k}(t \mid \mathcal{H}_t)$ and regularization terms to mitigate the overfitting issue. 




%% file: sections/related.tex
\textbf{Granger Causality.} Granger causality \citep{granger1969investigating} was initially developed for causal discovery in multivariate time series. Common approaches in Granger causality use linear models such as Vector Autoregressive Model (VAR) with a 
group lasso penalty \citep{arnold2007temporal,shojaie2010discovering}. Neural Granger causality \citep{tank2021neural}  relaxed the linearity assumption and introduced sparsity regularization in deep neural networks. \cite{lowe2022amortized} studied an amortized setting and proposed a deep variational model to handle time series with different underlying graphs. However, most of existing studies focus on time-series sampled at a regular time interval. For
event sequence data sampled at irregular time stamps,  two categories of prior works are directly relevant in Granger causality: Multivariate Hawkes process (MHP) and Neural Point Processes (NPP).

\paragraph{Multivariate Hawkes process.}
As suggested by Eq.~(\ref{eq:mhp_intensity}), where the kernel matrix depends only on the event types, maximum likelihood estimation with MHP only leads to type-level causal analysis~\citep{mei2017neural,xu2016learning,achab2017uncovering}. One approach to deriving instance-level causality is to leverage the MM algorithm~\citep{veen2008estimation,lewis2011nonparametric,li2015energy,wang2017human,Ide21NeurIPS}, where the instance-level causal strength is defined through a variational distribution of a lower bound of the log likelihood function. One limitation of this approach is that the instance-level causality is defined only through the first term of Eq.~(\ref{eq:mhp_intensity}). This is reminiscent of Cox's partial likelihood approach~\citep{cox1975partial} for the proportional hazard model, and hence, can be viewed as a heuristic. Another limitation is the linearity assumption in the intensity function. In complex domains, where nonlinear causal effects such as synergistic effects may exist, the restrictive parametric form can lead to a subnormal fit to the data.


\paragraph{Neural Point Processes.} NPPs combine point processes with deep neural networks. There are mainly two approaches for NPP. The first approach is based on recurrent neural networks (RNNs), while the other leverages the transformer architecture. In both approaches, the intensity function is typically represented as $\lambda_{k_{i}}(t_{i}\mid \bmh_{i-1})$ with $\bmh_{i-1}$ being the embedding vector of the event history $\{(t_j, k_j)\}^{i-1}_{j=1}$~\citep{xiao2017modeling}. 
The main advantage of the NPP-based approach is that neural networks, as universal sequence approximators, eliminate the need for carefully choosing a specific parametric form for the intensity function. However, they lose reference to individual event instances because of the embedding of $\calH_{t_i}$ into $\bmh_{i-1}$. As a result, retrieving instance-level dependencies generally requires additional model assumptions. One approach is to use the self-attention weights as a proxy of causal strength~\citep{xiao2019learning,zuo2020transformer,zhang2020self}, and the other is to introduce an ad-hoc dependency score defined independently of the maximum likelihood framework~\citep{zhang2020cause}. In both cases, it is not clear how those scores are related to the notion of Granger causality. 

While we also employ the transformer architecture, the key difference from the existing works is that our model is designed to \textit{maintain the additive structure} in the intensity function. This guarantees Granger-causality  in computing instance-level causal strengths, unlike the self-attention weights in the existing transformer Hawkes models. To the best of our knowledge, \ours{} is the first neural point process model that allows direct instance-level causal analysis.




%% file: sections/method_new.tex
In this section we present our \textbf{I}nstance-wise \textbf{S}elf-\textbf{A}ttentive \textbf{H}awkes \textbf{P}rocesses (\ours{}) model.

\textbf{Notation.}  Hereafter, vectors and matrices are denoted in bold italic (such as $\bmv$) and sans serif (such as $\sfW_V$), respectively. We use $^\top$ to denote the transpose of vectors and matrices. All vectors are column vectors. $\mathbb{R}$ and $\mathbb{R}_+$ denote the set of real numbers and non-negative real numbers, respectively. 

\subsection{Intensity Function}

\ours{} has two key features. First, it maintains the additive structure over the historical events in the intensity function, similar to MHP. Therefore, \ours{} inherits the interpretability of MHP for Granger causality. 
Second, \ours{} adopts an instance-aware parameterization of the kernel function. Specifically, we associate each event with a latent embedding vector $\bmx = g(t, k)$ and define the embedding function $g(t, k)$ as: 
\[g(t, k) =\mathrm{MLP}[t-t_i, \mathrm{MLP}(\bf{k})]\]
with the event type (as a $K$-dimensional one-hot vector $\bf{k}$) and the time difference (as $t_i - t_{i-1}$) for $\bmx_i$. We use an MLP layer to embed the one-hot vector for event type and concatenate it with the time difference to form a $M$-dimensional embedding vector. 

We assume the intensity function for an event embedding $\bmx$ in the form of:
\begin{equation}
    \label{eq:ISAHP_intensity_new}
    \lambda(\bmx \mid \calH_t)=
    \mu(\bmx \mid \calH_t) +  \sum_{j: t_j < t} \alpha(\bmx,\bmx_j) \phi(t-t_j\mid \bmx,\bmx_j),
\end{equation}
where $\mu$ is the background intensity,
 the function $\alpha(\bmx,\bmx_j) \in \mathbb{R^+}$ is the kernel function. 
 The kernel function characterizes the instance-level causal influence between events, generalizing the vanilla MHP, whose kernel matrix depends only on event types. 
%
The decay distribution $\phi(t-t_j\mid \bmx,\bmx_j)$ models the time decay of causal influence. In general, $\phi$ can be any distribution
depending on the statistical nature of the training dataset $\calD$. In our experiments, we assume a ``neural exponential distribution'':
\begin{gather}
    \label{eq:ISAHP_decay_new}
    \phi(t-t_j\mid \bmx,\bmx_j) = \gamma(\bmx, \bmx_j)\rme^{-\gamma(\bmx, \bmx_j)(t - t_j)},
\end{gather}
where $\gamma(\bmx, \bmx_j)$ is  the decay rate function.  
We use neural networks to model the functions $\gamma(\bmx, \bmx_j)$, $\mu(\bmx \mid \calH_t)$ and $\alpha(\bmx,\bmx_j)$, which will be discussed in later sections.

In the Hawkes-type model, event occurrence probability consists of two components: The spontaneous effect (the $\mu$ term) and the causal effects (the $\alpha$ term). While it is true that $\mu$ includes the elements of $A$ in average/aggregation, the individual causal effects are best captured by explicitly including the $\alpha$'s. Regularized MLE further resolves this issue near-optimally (c.f.\ \citep{eichler2017graphical}) and, hence, admits the reasonable interpretation that $\alpha_{\bmx,\bmx_j} = 0$ indicates Granger non-causality at the instance level.

\subsection{Self-attentive Architecture}

\begin{figure*}[h!]
    \centering
    \includegraphics[width=0.84\linewidth]{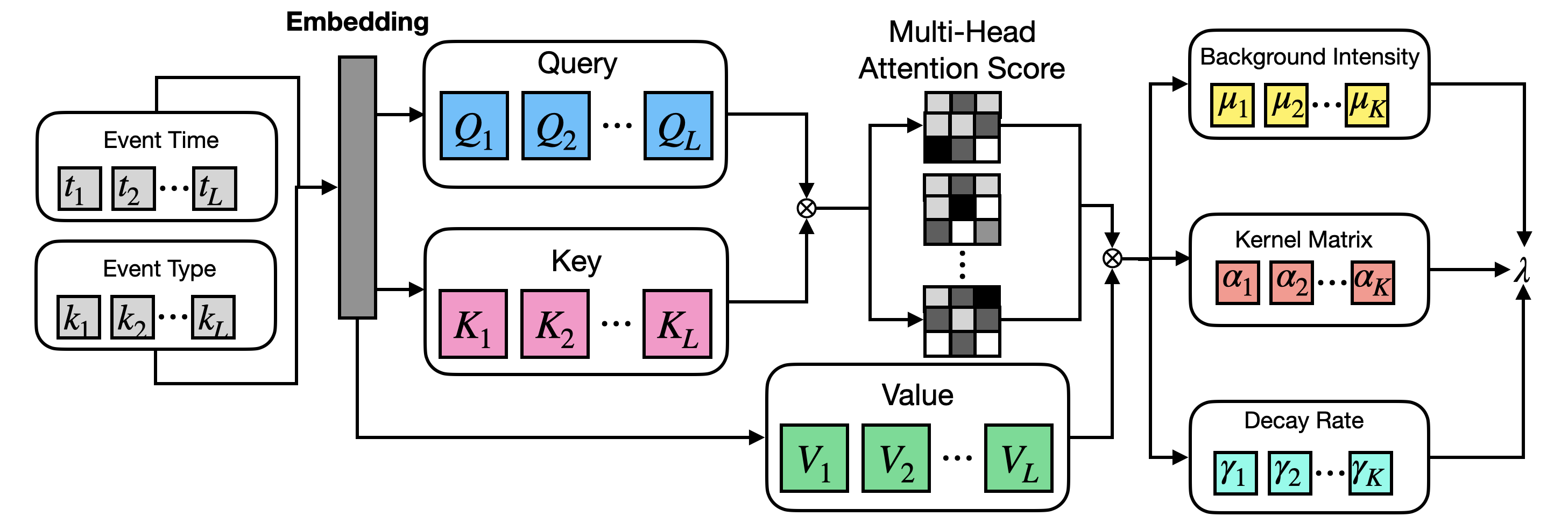}
    \vspace{-4mm}
    \caption{Instance-wise Self-attentive Hawkes Processes (\ours{}) architecture. The input is an event sequence identified by timestamp and event type. the output is an intensity function parameterized by background intensity, kernel matrix, and decay rate. 
    }
    \label{fig:ishap}
\end{figure*}

In multi-type event sequences, there may exist short-term temporal dependencies that the classical linear Hawkes models can capture, or non-trivial long-range temporal dependencies involving multiple event instances. To capture such dependencies, we introduce a neural architecture based on self-attention \cite{vaswani2017attention} to parametrize Eq.~(\ref{eq:ISAHP_intensity_new}). Fig.~\ref{fig:ishap} illustrates the model architecture. The embedding approach follows the standard key-value-query formalism of the transformer.

\paragraph{Self-Attention.} The embedding vectors $\{\bmx_j\}_{j=1}^L$ are linearly transformed to be the ``value'' vector:
\begin{align}\label{eq:value_vector_def}
    \bmv_j = \sfW_V^\top \bmx_j, \quad \mbox{or} \quad \sfV = \sfW_V^\top  \sfX,
\end{align}
where $\sfX \triangleq [\bmx_1, \ldots, \bmx_L] \in \mathbb{R}^{M\times L}$ and $\sfV \triangleq [\bmv_1, \ldots, \bmv_L] \in \mathbb{R}^{M_V\times L}$. $\sfW_V \in \mathbb{R}^{M \times M_V}$ is learned from the data. 

We capture the dependency among the events through  ``self-attention'' $A(\bmx, \bmx_j)\in \mathbb{R}$, defined by
\begin{gather}
    A(\bmx, \bmx_j) =
        \frac{\exp\left( \bmx^\top \sfK \bmx_j\right)\mathbb{I}(t > t_j)
        }{
            \sum_{l: t_l <t} \exp\left( \bmx^\top \sfK \bmx_l\right)
        }, 
\end{gather}
where $t$ is the timestamp associated with $\bmx$, and $\mathbb{I}(t > t_j)$ is the indicator function that assumes the value 1 if the argument is true and 0 otherwise. $\sfK \in \mathbb{R}^{M \times M }$ is a learnable parameter matrix. Following the notation of the transformer, $\sfK$ corresponds to $\sfW_Q\sfW_K^\top$, where $\sfW_Q$ and $\sfW_K$ are the transformation matrix for the queries and keys. As shown in  Fig.~\ref{fig:ishap}, the self-attention weights are represented as an $L\times L$ matrix $\sfA = [A_{i,j}]$ during training with $A_{i,j} \triangleq A(\bmx_i, \bmx_j)$. For  $K$ types of events, we use multi-head attention with $K$ different heads. 


\subsection{Kernel matrix, background intensity, and decay rate functions}


Now we infer $\gamma(\bmx, \bmx_j)$, $\mu(\bmx \mid \calH_t)$ and $\alpha(\bmx,\bmx_j)$ in Eq.~(\ref{eq:ISAHP_intensity_new}) and Eq.~(\ref{eq:ISAHP_decay_new}) according to the self-attentive architecture. For the background intensity  function, we use the following form:
%
\begin{align}\label{eq:baseline_intensity_new}
 \mu(\bmx \mid \calH_t)  = 
    \Bar{\mu}_k + \sigma\left((\bmw_k^\mu)^\top  \sum_{j: t_j < t} A(\bmx,\bmx_j) \bmv_j \right ).
\end{align}
Here, $\Bar{\mu}_k$ is the background intensity, and  $k$ is the event type encoded in $\bmx$.
The second term represents instance-specific effects in the background intensity, where $\sigma(\cdot)$ denotes an activation function. In our implementation, we used the sigmoid function. 
This term represents the averaged effect of causal interactions among the events. $\bmw_k^\mu \in \mathbb{R}^{M_V}$ is learned from data.

For the impact and decay rate functions, we adopt the following instance-specific formulation: 
\begin{align}
    & \alpha(\bmx,\bmx_j) = \sigma_+\left(A(\bmx, \bmx_j) (\bmw_k^\alpha)^\top 
      \bmv_j \right),  \nonumber \\ 
    & \gamma(\bmx,\bmx_j) 
    = \sigma_+\left(A(\bmx, \bmx_j) (\bmw_k^\gamma)^\top 
      \bmv_j  + b_k^\gamma \right).
\end{align}
where $\sigma_+(\cdot)$ is the softplus function applied element-wise on the vector argument, and the parameter vectors and matrices $\{\bmw_k^\alpha,\bmw_k^\gamma, b_k^\gamma\}_{k=1}^K$ are learned from the data. The input of these MLPs is $A(\bmx,\bmx_j)\bmv_j$, which can be viewed as a relevant component of $\bmv_j$ in terms of the impact on the target event represented by $\bmx$. To see how this model generalizes the vanilla MHP, imagine that $\sigma_+(\cdot)$ were the identity function. Then we have $\alpha(\bmx,\bmx_j) \to A(\bmx, \bmx_j)\tilde{w}^\alpha_{k,j}$, 
where $  \tilde{w}^\alpha_{k,j} \triangleq (\bmw_k^\alpha)^\top \bmv_j
$. By construction, the attention weight $ A(\bmx, \bmx_j)$ represents the similarity between $\bmx$ and $\bmx_j$. On the other hand, $\tilde{w}^\alpha_{k,j}$ can be interpreted as the relevance of $\bmv_j$ to type-$k$ events computed through the vector inner product. Compared to MHP's $\alpha_{k,k_j}$, which depends only on the event types, \ours{} looks at events at a finer granularity by using the embedding vectors.

\subsection{Maximum Likelihood Estimation}

We learn Eq.~(\ref{eq:ISAHP_intensity_new}) based on maximum likelihood estimation. To present the final objective function, we restore the sequence index $s$ hereafter. The main outcome of the unsupervised causal discovery task is $\alpha^s_{i,j} \triangleq \alpha(\bmx^s_i,\bmx^s_j)$, which quantifies the instance-level causal influence of the $j$-th event on the $i$-th event. As judged from Eq.~(\ref{eq:ISAHP_intensity_new}), $\alpha^s_{i,j}=0$ meets the definition of Granger-non-causality.


As a side product, the type-level causal dependency, denoted by $\Bar{\alpha}_{k,k'}$, can be obtained as the average of instance-level causal influence: 
\begin{align}\label{eq:type_level_causal_influence_def}
    \Bar{\alpha}_{k,k'}\triangleq
    \frac{1}{N_{k,k}}\sum_{s=1}^S
\sum_{i=1}^{L_s}\sum_{j=0}^i \delta_{k^s_j,k}\delta_{k^s_i,k}\alpha_{i,j}^s.
\end{align}
where $\delta_{k^s_j,k}$ etc.~is Kronecker's delta that is 1 if $k^s_j=k$ and 0 otherwise. $N_{k,k'}$ is the total counts of the event type pair $(k,k')$ in the dataset, defined by
$   N_{k,k'} \triangleq \sum_{s=1}^S
\sum_{i=1}^{L_s}\sum_{j=0}^i \delta_{k^s_j,k}\delta_{k^s_i,k}.
$

The final loss function to be minimized is now given by
\begin{align} 
\label{eq:loglikelihood1}
\mathcal{L}=&\sum_{k=1}^K\sum_{k'=1}^K\left(\omega_1 |\Bar{\alpha}_{k,k'}| + \omega_2\sigma^2_{k,k'}\right) + \nonumber \\
& \sum^S_{s=1}\sum^{L_s}_{i=1}\! \left[[
 \int^{t^s_i}_{t^s_{i-1}}\!\!\!\!\!\rmd t^\prime  \lambda(t^\prime, \bmx^s_i \mid \calH_{t^s_i})]\!-\!\ln \lambda(t^s_i, \bmx^s_i\mid \calH_{t^s_i}) \right].
\end{align}
where $\omega_1,\omega_2$ are the regularization strengths treated as hyperparameters. In the first term of Eq.~(\ref{eq:loglikelihood1}), we have introduced regularization terms for numerical stability and consistency within the same event type pair. Specifically, the type-level regularization (TLR) term $\omega_1 |\Bar{\alpha}_{k,k'}|$ is L$_1$ regularization on the mean of $\alpha$s sharing the same event type pair. We also include the variance regularization term~\citep{namkoong2017variance,huang2020feature}, with $\sigma^2_{k,k'}$ defined as
\begin{gather}
\label{eq:alpha_variance}
\sigma^2_{k,k'} \triangleq  \frac{1}{N_{k,k}}
\sum_{s,i,j}
\delta_{k^s_j,k}\delta_{k^s_i,k}
(\alpha_{i,j}^s - \Bar{\alpha}_{k,k'})^2,
\end{gather}
to control the variability within the event instances of the same event type pair $(k,k')$. 
As we increase the hyperparameter $\omega_2$, \ours{} is encouraged to provide a generative process that is similar to the vanilla MHP for a given decay model.
The second term of  Eq.~(\ref{eq:loglikelihood1}) corresponds to the negative log-likelihood function, where we used the well-known relationship between the intensity function and the log-likelihood (See, e.g.,~\citep{daley2003introduction}). The integral can be performed analytically for the neural exponential distribution.

%


%% file: sections/exp.tex
We evaluate Granger causality inference at both the type level and the instance level. We aim to verify that (a) \ours{} outperforms other baselines for the type-level causality discovery task; (b) there is a positive correlation between performances in the type-level Granger causality inference and the instance-level event type prediction, allowing \ours{} to accurately predict the type of next event instance; (c) \ours{} can capture complex synergistic causal effects over multiple event types at the instance level. 

\subsection{Experimental Set-up}

\paragraph{Datasets.} For empirical validation, we used two datasets of different sizes: Synergy and Memetracker (MT). These two datasets were chosen since they contain non-linear causal interactions that are challenging for classical models hence motivating a new solution approach. Out of those benchmark datasets from \citep{zhang2020cause}, they are the only datasets with non-linear causality that require instance-level causality analysis. 
The details of the data set and statistics are summarized in Appendix \ref{subsec: statistics}.


\paragraph{Baselines.} We compared \ours{} with six baselines: Three belong to the category of the classical MHP and three from the NPP family. 

\begin{itemize}

 \item  \texttt{HExp}: MHP in Eq.~(\ref{eq:mhp_intensity}) with the exponential decay model.

 \item  \texttt{HSG}: MHP with a Gaussian mixture decay~\citep{xu2016learning}, which is known as the state-of-the-art parametric model for Granger causality in classical MHP.

\item  \texttt{CRHG}:
 A sparse Granger-causal learning framework based on a cardinality-regularized Hawkes process (CRHG) ~\citep{Ide21NeurIPS}. Note that CRHG is designed to learn from a single event sequence. To incorporate this baseline into type-level causality analysis, we concatenate sequences from the dataset to form a long sequence. 

 \item  \texttt{RPPN}:
Recurrent Point Process Networks (RPPN) ~\citep{xiao2019learning}. An RNN (recurrent neural network)-based NPP that supports Granger causality inference based on an added attention layer.

 \item  \texttt{SAHP}:
Self-Attentive Hawkes Process (SAHP) ~\citep{zhang2020self}. A transformer-based NPP that enables Granger causality analysis based on the attention mechanism. 
It directly uses the self-attention from its transformer architecture to aggregate the influence
from historical events in determining the intensity function for the next event. 


 \item  \texttt{CAUSE}:
Causality from attributions on sequence of events (CAUSE) ~\citep{zhang2020cause}. An RNN-based framework for inferring Granger causality. It includes a post-training step to infer the instance-level Granger causality using an attribution method called the integrated gradient. Note that \ours{} does not require any post-training step and can directly infer the instance-level Granger causality based on its additive intensity function.   

\end{itemize}
\paragraph{Evaluation Metrics} We conducted three different experiments to validate the proposed \ours{} in addition to an ablation study to validate TLR. While the main motivation of \ours{} is instance-level Granger causal analysis, we include proxy tasks involving type-level inference as well as instance-level event-type prediction, due to the scarcity of ground truth data on instance-level causality.

(1) Type-level Granger causal discovery. We used the area under the curve (AUC) of the true positive vs.~false positive curve and Kendall’s $\tau$ coefficient to measure the accuracy of the inferred Granger causality matrix compared to the ground truth. 
(2) Next event-type prediction. This can be reduced to a multi-class classification problem given its timestamp. We used the classification accuracy to measure the performance. (3) Instance-level causal discovery. 
We picked a representative sequence pair 
involving synergistic causal interactions to highlight qualitative differences from the baselines. We also conducted statistical analysis by measuring the ratio between synergistic and non-synergistic contribution scores. For (1) and (2), we follow the setting of \citep{zhang2020cause} and report the average results based on five-fold cross-validation. 

\paragraph{Implementation Details and Hyperparameter Configurations}

The \ours{} hyperparameter settings for Synergy and MT experiments are shown in Appendix \ref{subsec: implementation}. These optimal hyperparameter settings were selected based on five-fold cross-validation. We use the Adam optimizer for training. The implementation details for other baselines are in Appendix \ref{subsec: implementation}.

\begin{table*}[t]
\begin{center}
\caption{Results for Granger causality discovery on two datasets with ground-truth causality. For both AUC and Kendall’s $\tau$, larger values are better. The result shows that the proposed method, \ours{}, is the most accurate and robust overall.}
\label{tb:type_granger1}
\small
\begin{tabular}{l|c c| c c}
\toprule
 &               \multicolumn{2}{c|}{\textbf{Synergy}} &                 \multicolumn{2}{c}{\textbf{MT}}             \\ \midrule
       & AUC     & Kendall’s $\tau$    & AUC  & Kendall’s $\tau$    \\ \midrule
HExp   &        0.885 $\pm$ 0.014  & 0.361 $\pm$ 0.013        &  0.616 $\pm$ 0.021     &     0.061 $\pm$ 0.011               \\ \hline
HSG    &    0.306 $\pm$ 0.063    & -0.182 $\pm$ 0.059     &   0.705 $\pm$ 0.015   &    0.105 $\pm$ 0.007                \\ \hline
CRHG  &   0.515 $\pm$ 0.033  &    0.014 $\pm$ 0.031     &     0.611 $\pm$ 0.01     &     0.079 $\pm$ 0.006       \\ \hline
CAUSE  &   0.761 $\pm$ 0.068    & 0.244 $\pm$ 0.064        &  0.739 $\pm$ 0.042 & 0.127 $\pm$ 0.022     \\ \hline
RPPN   &   0.827 $\pm$ 0.017    & 0.307 $\pm$ 0.016        &  0.437 $\pm$ 0.008 & -0.031 $\pm$ 0.004     \\ \hline
SAHP  &  0.182 $\pm$ 0.119     &   -0.298 $\pm$ 0.112      & 0.832 $\pm$ 0.012  &  \bf{0.251} $\pm$ 0.016    \\ \hline
ISAHP  & \bf{0.967} $\pm$ 0.007   & \bf{0.438} $\pm$ 0.006       &  \bf{0.835} $\pm$ 0.002      &     0.247 $\pm$ 0.001 \\
\bottomrule
\end{tabular}
\end{center}
\end{table*}
\subsection{Experimental Results}

\paragraph{Type-level Causality Analysis} We evaluate the performance of type-level Granger causality inference. 
Table~\ref{tb:type_granger1} exhibits the accuracy measures for Granger causality inference using AUC and Kendall's $\tau$ for \ours{} as well as 6 baselines. 
We see that \ours{} generally outperforms all baselines. For AUC, it is the best among all methods. For Kendall's $\tau$, it is the best for the Synergy dataset and is almost tied with the best method (SAHP) for the MT dataset. Note that \ours{} always has the smallest variance,
indicating that \ours{} is the most robust.

\begin{table}[htbp]
\begin{center}
\caption{Prediction accuracy in next-event-type prediction. The higher, the better. \ours{} outperforms all the baselines.}
\small
\label{tb:pred_acc}
\begin{tabular}{l|l|l}
\toprule
 &         \textbf{Synergy} &               \textbf{MT}            \\ \midrule
HExp   &        0.349 $\pm$ 0.013       &  0.862 $\pm$ 0.004          \\ \hline
HSG    &      0.364 $\pm$ 0.009     &     0.835 $\pm$ 0.006              \\ \hline
CAUSE  &      0.37 $\pm$ 0.012       &   0.905 $\pm$ 0.004    \\ \hline
RPPN   &         0.364 $\pm$ 0.01   &  0.748 $\pm$ 0.056  \\ \hline
SAHP  &    0.343 $\pm$ 0.013       &  0.459 $\pm$ 0.033      \\ \hline
\ours{}  &    \bf{0.471} $\pm$ 0.008      &   \bf{0.974} $\pm$ 0.002       \\ 
\bottomrule
\end{tabular}
\end{center}
\end{table}

For the MHP baselines ({HExp}, {HSG}, {CRHG}), we see that they perform poorly for both Synergy and MT. This is expected to some extent as Synergy involves synergistic effects between multiple causes and MT is based on a real-world dataset including non-linear effects. The underlying data generation mechanisms do not adhere to the linearity assumption of the (type level) intensity functions of these models.

For the two NPP baselines (RPPN, SAHP) that use the self-attention weights for (pseudo) causal attribution, we observe that their performance is quite unstable. SAHP reaches the start-of-the-art performance on MT dataset for Kendall's $\tau$ and is the second best on AUC, but performs the worst on the Synergy dataset. Similarly, RPPN has a relatively good performance on Synergy but is the worst on the MT dataset. These results indicate that using attention as attribution can be unstable depending on the data characteristics and there is no guarantee on the performance. One key issue is that they do not directly use the self-attention to parameterize the intensity function. Instead, they perform matrix multiplication between the self-attention scores and the value tensor. This step fuses information from the historical events and masks the pairwise causal relationships at the instance level. Similar results have been observed and studied recently \citep{serrano2019attention, jain2019attention}.
This contrasts with our approach that directly uses the attention scores to parameterize the intensity function, which is one of our key contributions. 

Note that CAUSE is the second most robust baseline. However, it requires additional computational overhead with the post-training step which makes $O(SK/B)$ invocations of a rather expensive attribution procedure, as we discussed earlier.

\paragraph{Instance-Level Event Type Prediction} We consider the event type prediction at the instance level. 

In Table \ref{tb:pred_acc}, we compare all the methods in terms of the classification accuracy score. It is clear that \ours{} performs better than all baselines on both datasets. Specifically, \ours{} reaches $27.3\%$ relative improvement over the second-best method on the Synergy dataset and $7.62\%$ relative improvement on the MT dataset.

Another interesting finding is that although SAHP performs well on the type-level Granger causality discovery for the MT dataset, its event-type prediction accuracy for the same dataset is the worst. A similar phenomenon is observed for RPPN on the Synergy dataset. This is another indication that naively using attention for causal attribution can be unstable. 

\begin{figure*}[t]
    \centering
    \includegraphics[width=0.95\linewidth]{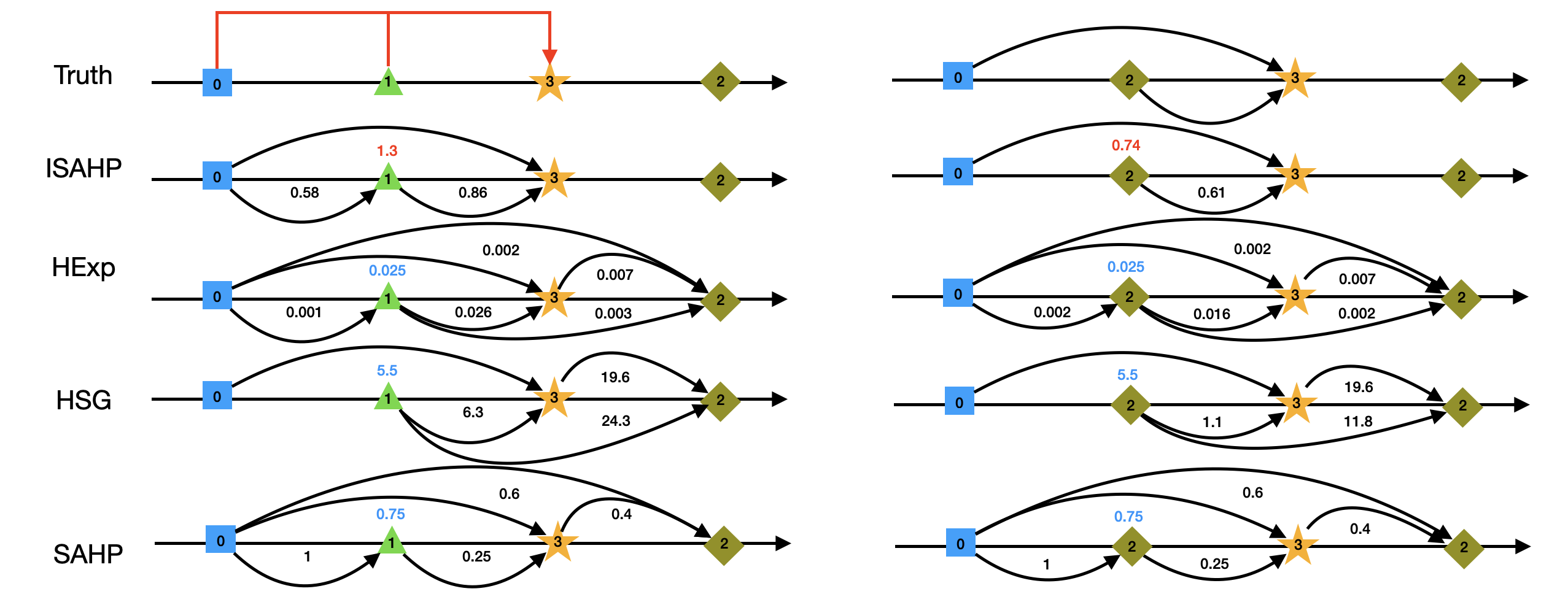}
    \caption{Instance-level causality analysis. The weight of the edge from the first event to the third is what we compare for the synergistic (left) and non-synergistic (right) sequences. Red numbers represent successful cases and blue numbers represent failure cases. \ours{} is the only one that successfully captures the synergistic effect at the instance level.}
    \label{fig:ishap_instance}
\end{figure*}

\paragraph{Instance-level Causality Analysis} One of the key advantages of \ours{} is that it can perform accurate {\em instance-level} causality analysis. Here we present anecdotal evidence of this characteristic, together with statistical analysis. Fig \ref{fig:ishap_instance} top exhibits two similar event sequences we sampled from the Synergy dataset. Each of them has four events on the timeline. Each event has been assigned a numerical label indicating its event type. 
The first and second events of the first sequence (on the left) have a synergistic effect on the third event (as indicated by the red square arrow). 
In contrast, in the second sequence (on the right), the first and second events have causal relationships with the third event, but independently (indicated by the black arrows). 
To be more precise, the ground truth PGEM model used to generate the data contains a type-level causal relationship $(0 \land 1) \rightarrow 3$ but not $(0 \land 2) \rightarrow 3$. 

\begin{table}[htbp]
\begin{center}
\caption{Averaged ratio between synergistic and non-synergistic instance-level contribution score on sequences with patterns $'0\#32'$, $'0\#43'$, and $'0\#23'$. A higher ratio means better performance.} \label{tb:instance}
\small
\begin{tabular}{l|l|l|l}
\toprule
             & '0\#32' & '0\#43' & '0\#23' \\ \midrule
HExp &  1 & 1    & 1    \\ \hline
HSG &  1 & 1    & 1    \\ \hline
SAHP & 0.997 & 0.994 & 1.025 \\ \hline
\ours{} & \bf{1.253} & \bf{1.155}   & \bf{1.151}   \\ 
\bottomrule
\end{tabular}
\end{center}
\end{table}

\begin{table*}[h]
\begin{center}
\caption{Ablation study on type-level regularization.}\label{tb:ablation}
\small
\begin{tabular}{l|c|c|c}
\toprule
&  \multicolumn{3}{c}{\textbf{Synergy}}       \\ \hline
  TLR      & ACC  & AUC     & Kendall’s $\tau$    \\ \hline
 \xmark  &     0.344 $\pm$ 0.011       &              0.903 $\pm$ 0.127    &    0.378 $\pm$ 0.119         \\ \hline

 \checkmark  &     \bf{0.471} $\pm$ 0.008     &         \bf{0.967} $\pm$ 0.007   & \bf{0.438} $\pm$ 0.006  \\  
 \midrule
  &  \multicolumn{3}{c}{\textbf{MT}}     \\ \hline
 \xmark & 0.95 $\pm$ 0.007   &    0.814$\pm$0.005      &  0.232$\pm$0.004 \\ \hline
 \checkmark & \bf{0.974} $\pm$ 0.002  & \bf{0.835} $\pm$ 0.002      &     \bf{0.247} $\pm$ 0.001     \\
 \bottomrule
\end{tabular}
\end{center}
\end{table*}

We would expect an effective causal attribution method to differentiate between the contribution from the first event to the third event under the synergistic and non-synergistic contexts. Fig \ref{fig:ishap_instance} shows that \ours{} successfully assigns larger contribution scores in the synergistic case: The weight of the edge (from the first event to the third) is $1.3$ for the synergistic case on the left, while the edge weight for the non-synergistic case on the right is $0.74$. 
On the other hand, all $3$ baselines, Hexp, HSG, and SAHP fail.
Hexp and HSG are not able to capture the synergistic effects (they assign identical weights in both cases) because they are parameterized to infer Granger causality at the type level. 
SAHP is intended to infer Granger causality at the instance level, but assigns essentially identical weights in the two cases.

To further verify the superior performance of \ours, we perform statistical analysis by traversing the dataset to identify all sub-sequences that match the patterns $'0\#32'$, $'0\#43'$, and $'0\#23'$, where event type $\# \in \{0,1,2,3,4\}$. The synergistic effect occurs only when $\# = 1$. Table \ref{tb:instance} shows the ratio of the average inferred instance-level contribution of events with type $0$ to events with type $3$ in the presence and absence of a synergistic effect. The results show that \ours{} always has the optimal performance compared with the baselines. \ours{} is able to achieve that because of the tight coupling between the type level and instance level causal learning. In effect, \ours{} captures synergistic effects at the type level using its additive structure at the instance level.




\paragraph{Ablation Study} Finally, we conducted an ablation study on type-level regularization (TLR) to verify that including TLR does improve ISHAP's performance. For each dataset, we compared TLR and non-TLR cases for both type-level causality analysis and type prediction, based on AUC, Kendall's $\tau$, and accuracy (ACC). The results in Table \ref{tb:ablation} show that including TLR does improve the model performance for both datasets.




%% file: sections/conc.tex

We address the problem of  discovering instance-level
causal structures from asynchronous, interdependent,
multi-type event sequences in an unsupervised manner.
We proposed a novel self-attentive deep Hawkes model, which enjoys the ability of instance-level causal discovery and the high expressiveness of deep Hawkes models. It is the first neural point process model that can be directly used for instance-level causal discovery to the best of our knowledge. Our empirical evaluation showed that the proposed instance-aware model significantly improved the performance of type-level tasks as well, suggesting that instance- and type-level causal inference tasks are tightly coupled.
One of the important future research topics is to conduct further empirical validation of its performance in instance-level causal analysis, while addressing the challenge due to the scarcity of instance-level ground truth data.

%% file: sections/appendix.tex

\subsection{Time Complexity Analysis}
\label{subsec: complexity}
ISAHP can directly infer Granger causality at the event instance level through a forward pass during the evaluation phases. Compared with NPP benchmark CAUSE, ISAHP no longer requires a post-training process, which would take $O(SK/B)$ invocations of the (integrated gradient) attribution process, where $S$ is the number of sequences, $K$ is the number of event types, and $B$ is the batch size. ISAHP avoids this computation which can be taxing when the number of sequences is large.

\subsection{Statistics of datasets}
\label{subsec: statistics}

\texttt{Synergy}: A synthetic dataset with complex type-level interaction including a synergistic causal dependency triggered by a pair of event types on the third event type. We used it to test whether ISAHP can reproduce such complex type-level interactions through its pairwise kernel function. The data is generated by a proximal graphical event model (PGEM) \citep{bhattacharjya2018proximal}. The ground-truth causality matrix is binary and based on the dependency graph of the PGEM simulator.

\texttt{MemeTracker (MT)}: A larger-scale real-world dataset, where each sequence represents how a phrase or quote appeared on various online websites over time. We chose the top $25$ websites as our event types from August 2008. The ground-truth causality matrix is weighted and approximated by whether one site contains any URL links to another site \citep{achab2017uncovering,xiao2019learning}. 

The dataset statistics are summarized in Table \ref{tb:statistics}, where $S$ is the number of sequences, $K$ is the number of event types, $L_s$ is the sequence length for each sequence $s$.

\begin{table}[h]
\begin{center}
\caption{Statistics of datasets used.}
\label{tb:statistics}
\footnotesize
\begin{tabular}{ l|c|c|c|c }
\toprule
Dataset    & $S$     & $K$  & $\sum_{s=1}^S L_s$     & Ground Truth \\ \hline
\textbf{Synergy}    & 1,000 & 5  & 16,101  & Binary       \\ \hline
\textbf{MT}         & 8,703 & 25 & 90,787 & Binary     \\ \bottomrule
\end{tabular}
\end{center} 
\end{table}

\subsection{Implementation Details and Hyperparameter Configurations}
\label{subsec: implementation}
For MHP baselines (HExp, HSG), we use the implementations provided by the tick package \citep{bacry2017tick}. For RPPN and CAUSE, the implementation is from \citep{zhang2020cause}. For SAHP,  we use the implementation by \citep{zhang2020self}. The hyperparameters for these baselines were tuned by cross-validation. Specifically, we tuned the learning rate, batch size, and hidden size. The tuned hyperparameter configurations for \ours{} are shown in Table \ref{tb:hyper}.

\begin{table}[h]
\begin{center}
\caption{Hyperparameter configurations for ISAHP.}
\label{tb:hyper}
\begin{tabular}{ l|c|c}
\toprule
Hyperparameter    & \textbf{Synergy}    & \textbf{MT}   \\ \hline
Learning rate   & $0.001$ & $0.001$       \\ \hline
Batch Size   & $8$ &   $16$    \\ \hline
Hidden Size   & $10$ &  $50$       \\ \hline
Number of Attention Heads   & $2$ & $2$        \\ \hline
$\omega_1$   & $0.025$ &   $0.$     \\ \hline
$\omega_2$   & $0.25$ &   $5.$     \\ \bottomrule
\end{tabular}
\end{center} 
\end{table}

